\documentclass{article}
\usepackage{spconf, amsmath, amsfonts, graphicx}
\usepackage{url, color, subfigure}
\usepackage[colorlinks=true,urlcolor=black,linkcolor=black,citecolor=black]{hyperref}
\usepackage{algorithm}
\usepackage[noend]{algorithmic}
\usepackage{xfrac}

\usepackage{enumitem}
\setlist{leftmargin=*} 

\title{High-dimensional sequence transduction}
\name{Nicolas Boulanger-Lewandowski \qquad Yoshua Bengio \qquad Pascal Vincent\thanks{The authors would like to thank NSERC, CIFAR and the Canada Research Chairs for funding, and Compute Canada/Calcul Qu\'ebec for computing resources.}}
\address{Dept. IRO, Universit\'e de Montr\'eal\\Montr\'eal (QC), H3C 3J7, Canada}

\begin{document}
\ninept
\maketitle
\begin{abstract}
We investigate the problem of transforming an input sequence into a high-dimensional output sequence in order to transcribe polyphonic audio music into symbolic notation.
We introduce a probabilistic model based on a recurrent neural network that is able to learn realistic output distributions given the input and we devise an efficient algorithm to search for the global mode of that distribution.
The resulting method produces musically plausible transcriptions even under high levels of noise and drastically outperforms previous state-of-the-art approaches on five datasets of synthesized sounds and real recordings, approximately {\em halving the test error rate}.
\end{abstract}
\section{Introduction}

Machine learning tasks can often be formulated as the transformation, or \emph{transduction}, of an input sequence into an output sequence: speech recognition, machine translation, chord recognition or automatic music transcription, for example.
Recurrent neural networks (RNN)~\cite{rumelhart1986} offer an interesting route for sequence transduction~\cite{graves2012sequence} because of their ability to represent arbitrary output distributions involving complex temporal dependencies at different time scales.

When the output predictions are high-dimensional vectors, such as tuples of notes in musical scores, it becomes very expensive to enumerate all possible configurations at each time step.
One possible approach is to capture high-order interactions between output variables using restricted Boltzmann machines (RBM)~\cite{Smolensky1986} or a tractable variant called NADE~\cite{larochelle2011nade}, a weight-sharing form of the architecture introduced in~\cite{Bengio+Bengio-NIPS99}.
In a recently developed probabilistic model called the RNN-RBM, a series of distribution estimators (one at each time step) are conditioned on the deterministic output of an RNN~\cite{boulanger2012modeling, sutskever2008recurrent}.
In this work, we introduce an input/output extension of the RNN-RBM that can learn to map input sequences to output sequences, whereas the original RNN-RBM only learns the output sequence distribution.
In contrast to the approach of \cite{graves2012sequence} designed for discrete output symbols, or one-hot vectors, our high-dimensional paradigm requires a more elaborate inference procedure.
Other differences include our use of second-order Hessian-free (HF)~\cite{martens2011learning} optimization\footnote{Our code is available online at \url{http://www-etud.iro.umontreal.ca/~boulanni/icassp2013}.} but not of LSTM cells~\cite{hochreiter1997long} and, for simplicity and performance reasons, our use of a single recurrent network to perform both transcription and temporal smoothing.
We also do not need special ``null" symbols since the sequences are already aligned in our main task of interest: polyphonic music transcription.

The objective of polyphonic transcription is to obtain the underlying notes of a polyphonic audio signal as a symbolic \emph{piano-roll}, i.e. as a binary matrix specifying precisely which notes occur at each time step.
We will show that our transduction algorithm produces more musically plausible transcriptions in both noisy and normal conditions and achieve superior overall accuracy~\cite{bay2009evaluation} compared to existing methods.
Our approach is also an improvement over the hybrid method in \cite{boulanger2012modeling} that combines symbolic and acoustic models by a product of experts and a greedy chronological search, and \cite{cemgil2006generative} that operates in the time domain under Markovian assumptions.
Finally, \cite{bock2012polyphonic} employs a bidirectional RNN without temporal smoothing and with independent output note probabilities.
Other tasks that can be addressed by our transduction framework include automatic accompaniment, melody harmonization and audio music denoising.


\section{Proposed architecture} \label{sec:architecture}

\subsection{Restricted Boltzmann machines} \label{sec:rbm}

An RBM is an energy-based model where the joint probability of a given configuration of the visible vector~$v\in\{0,1\}^N$ (output) and the hidden vector~$h$ is:
\begin{equation} \label{eq-binrbm}
P(v,h) = \exp({-b_v^\mathrm Tv - b_h^\mathrm Th -h^\mathrm TWv})/Z
\end{equation}
where $b_v$, $b_h$ and $W$ are the model parameters and $Z$ is the usually intractable partition function.
The marginalized probability of $v$ is related to the free-energy $F(v)$ by $P(v)\equiv e^{-F(v)}/Z$:
\begin{equation} \label{eq-fe}	
F(v) = -b_v^\mathrm Tv - \sum_i \log(1+e^{b_h+Wv})_i
\end{equation}
The gradient of the negative log-likelihood of an observed vector $v$ involves two opposing terms, called the positive and negative phase:
\begin{equation}
\frac{\partial (-\log P(v))}{\partial\Theta} = \frac{\partial F(v)}{\partial\Theta} - \frac{\partial (-\log Z)}{\partial\Theta}
\end{equation}
where $\Theta\equiv\{b_v, b_h, W\}$.
The second term can be estimated by a single sample $v^*$ obtained from a Gibbs chain starting at $v$:
\begin{equation} \label{eq-cost}
\frac{\partial (-\log P(v))}{\partial\Theta} \simeq \frac{\partial F(v)}{\partial\Theta} - \frac{\partial F(v^*)}{\partial\Theta}.
\end{equation}
resulting in the well-known contrastive divergence algorithm~\cite{hinton2002training}.

\subsection{NADE}

The neural autoregressive distribution estimator (NADE) \cite{larochelle2011nade} is a tractable model inspired by the RBM.
NADE is similar to a fully visible sigmoid belief network in that the conditional probability distribution of a visible unit $v_j$ is expressed as a nonlinear function of the vector $v_{<j} \equiv \{v_k, \forall k<j\}$:
\begin{equation} \label{eq-pnade}
P(v_j=1|v_{<j}) = \sigma(W^\top_{:,j}h_j + (b_v)_j)
\end{equation}
\begin{equation}
h_j = \sigma(W_{:,<j}v_{<j} + b_h)
\end{equation}
where $\sigma(x)\equiv(1+e^{-x})^{-1}$ is the logistic sigmoid function.

In the following discussion, one can substitute RBMs with NADEs by replacing equation~(\ref{eq-cost}) with the exact gradient of the negative log-likelihood cost $C\equiv -\log P(v)$:
\begin{equation} \label{eq:nade_gbv}
\frac{\partial C}{\partial (b_v)_j} = P(v_j=1|v_{<j}) - v_j
\end{equation}
\begin{equation} \label{eq:nade_gbh}
\frac{\partial C}{\partial b_h} = \sum_{k=1}^N \frac{\partial C}{\partial (b_v)_k} W_{:,k} h_k(1-h_k)
\end{equation}
\begin{equation} \label{eq:nade_gw}
\frac{\partial C}{\partial W_{:,j}} = \frac{\partial C}{\partial (b_v)_j} h_j + v_j \sum_{k=j+1}^N \frac{\partial C}{\partial (b_v)_k} W_{:,k} h_k(1-h_k)
\end{equation}
In addition to the possibility of using HF for training, a tractable distribution estimator is necessary to compare the probabilities of different output sequences during inference.

\subsection{The input/output RNN-RBM} \label{sec:io_rnnrbm}

\begin{figure}
\centering
\def\svgwidth{0.9\columnwidth}
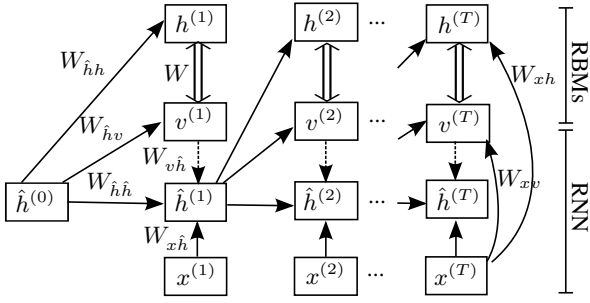
\vskip -2pt
\caption{Graphical structure of the I/O RNN-RBM.
Single arrows represent a deterministic function, double arrows represent the hidden-visible connections of an RBM,
dotted arrows represent optional connections for temporal smoothing.
The $x\rightarrow\{v,h\}$ connections have been omitted for clarity at each time step except the last.
}
\label{fig:model}
\end{figure}

The I/O RNN-RBM is a sequence of conditional RBMs (one at each time step) whose parameters $b_v^{(t)}, b_h^{(t)}, W^{(t)}$ are time-dependent and depend on the sequence history at time $t$, denoted $\mathcal A^{(t)}\equiv\{x^{(\tau)},v^{(\tau)} | \tau<t\}$ where $\{x^{(t)}\},\{v^{(t)}\}$ are respectively the input and output sequences.
Its graphical structure is depicted in Figure~\ref{fig:model}.
Note that by ignoring the input $x$, this model would reduce to the RNN-RBM~\cite{boulanger2012modeling}.
The I/O RNN-RBM is formally defined by its joint probability distribution:
\begin{equation} \label{eq-jpd}
P(\{v^{(t)}\}) = \prod_{t=1}^T P(v^{(t)}|\mathcal A^{(t)})
\end{equation}
where the right-hand side multiplicand is the marginalized probability of the $t^\mathrm{th}$ RBM (eq.~\ref{eq-fe}) or NADE (eq.~\ref{eq-pnade}).

Following our previous work, we will consider the case where only the biases are variable:
\begin{equation} \label{eq-bh}
b_h^{(t)} = b_h + W_{\hat hh}\hat h^{(t-1)} + W_{xh}x^{(t)}
\end{equation}
\begin{equation} \label{eq-bv}
b_v^{(t)} = b_v + W_{\hat hv}\hat h^{(t-1)} + W_{xv}x^{(t)}
\end{equation}
where $\hat h^{(t)}$ are the hidden units of a single-layer RNN:
\begin{equation} \label{eq-inf}
\hat h^{(t)} =\sigma( W_{v\hat h} v^{(t)} + W_{\hat h\hat h}\hat h^{(t-1)} + W_{x\hat h}x^{(t)} + b_{\hat h} )
\end{equation}
where the indices of weight matrices and bias vectors have obvious meanings.
The special case $W_{v\hat h}=0$ gives rise to a transcription network without temporal smoothing.
Gradient evaluation is based on the following general scheme:
\begin{enumerate}
\item Propagate the current values of the hidden units $\hat h^{(t)}$ in the RNN portion of the graph using (\ref{eq-inf}),
\item Calculate the RBM or NADE parameters that depend on $\hat h^{(t)},x^{(t)}$ (eq.~\ref{eq-bh}-\ref{eq-bv}) and obtain the log-likelihood gradient with respect to $W$, $b_v^{(t)}$ and $b_h^{(t)}$ (eq.~\ref{eq-cost} or eq.~\ref{eq:nade_gbv}-\ref{eq:nade_gw}),
\item Propagate the estimated gradient with respect to $b_v^{(t)}, b_h^{(t)}$ backward through time (BPTT)~\cite{rumelhart1986} to obtain the estimated gradient with respect to the RNN parameters.
\end{enumerate}

By setting $W=0$, the I/O-RNN-RBM reduces to a regular RNN that can be trained with the cross-entropy cost:
\begin{equation} \label{eq:rnn_ce}
L(\{v^{(t)}\}) = \frac1T \sum_{t=1}^T \sum_{j=1}^N -v_j^{(t)} \log p_j^{(t)} - (1-v_j^{(t)}) \log (1-p_j^{(t)})
\end{equation}
where $p^{(t)} = \sigma(b_v^{(t)})$ and equations~(\ref{eq-bv}) and (\ref{eq-inf}) hold.
We will use this model as one of our baselines for comparison.

A potential difficulty with this training scenario stems from the fact that since $v$ is known during training, the model might (understandably) assign more weight to the symbolic information than the acoustic information.
This form of \emph{teacher forcing} during training could have dangerous consequences at test time, where the model is autonomous and may not be able to recover from past mistakes.
The extent of this condition obviously depends on the ambiguousness of the audio and the intrinsic predictability of the output sequences, and can also be controlled by introducing noise to either $x^{(\tau)}$ or $v^{(\tau)},\tau<t$, or by adding the regularization terms $\alpha (|W_{xv}|^2+|W_{xh}|^2) + \beta(|W_{\hat hv}|^2+|W_{\hat hh}|^2)$ to the objective function. 
It is trivial to revise the stochastic gradient descent updates to take those penalties into account.

\vspace{-5pt}
\section{Inference} \label{sec:inference}
\vspace{-2pt}

A distinctive feature of our architecture are the (optional) connections $v\rightarrow\hat h$ that implicitly tie $v^{(t)}$ to its history $\mathcal A^{(t)}$ and encourage coherence between successive output frames, and temporal smoothing in particular.
At test time, predicting one time step $v^{(t)}$ requires the knowledge of the previous 
decisions on $v^{(\tau)}$ (for $\tau<t$) which are yet uncertain (not chosen optimally), and proceeding in a greedy chronological manner does not necessarily yield configurations that maximize the likelihood of the complete sequence\footnote{Note that without temporal smoothing ($W_{v\hat h}=0$), the $v^{(t)},1\le t\le T$ would be conditionally independent given $x$ and the prediction could simply be obtained separately at each time step $t$.}.
We rather favor a global search approach analogous to the Viterbi algorithm for discrete-state HMMs.
Since in the general case the partition function of the $t^\mathrm{th}$~RBM depends on $\mathcal A^{(t)}$, comparing sequence likelihoods becomes intractable, hence our use of the tractable NADE. 

Our algorithm is a variant of beam search for high-dimensional sequences, with beam width $w$ and maximal branching factor $K$ (Algorithm~\ref{alg-beamsearch}).
Beam search is a breadth-first tree search where only the $w$ most promising paths (or nodes) at depth $t$ are kept for future examination.
In our case, a node at depth $t$ corresponds to a subsequence of length $t$, and all descendants of that node are assumed to share the same sequence history $\mathcal A^{(t+1)}$; consequently, only $v^{(t)}$ is allowed to change among siblings.
This structure facilitates identifying the most promising paths by their cumulative log-likelihood.
For high-dimensional output however, any non-leaf node has exponentially many children ($2^N$), which in practice limits the exploration to a fixed number $K$ of siblings.
This is necessary because enumerating the configurations at a given time step by decreasing likelihood is intractable (e.g. for RBM or NADE) and we must resort to stochastic search to form a pool of promising children at each node. 
Stochastic search consists in drawing $S$ samples of $v^{(t)}|A^{(t)}$ and keeping the $K$ unique most probable configurations.
This procedure usually converges rapidly with $S\simeq10K$ samples, especially with strong biases coming from the conditional terms.
Note that $w=1$ or $K=1$ reduces to a greedy search, and $w=2^{NT},K=2^N$ corresponds to an exhaustive breadth-first search.

\begin{algorithm}[t]
\caption{\sc High-dimensional beam search}
Find the most likely sequence $\{v^{(t)},1\le t\le T\}$ under a model $m$ with beam width $w$ and branching factor $K$.

\begin{algorithmic}[1] \label{alg-beamsearch}
\STATE $q \leftarrow$ min-priority queue
\STATE $q$.insert($0,m$)
\FOR {$t=1\ldots T$}
  \STATE $q' \leftarrow$ min-priority queue of capacity $w$~$^\star$
  \WHILE{$l,m \leftarrow q$.pop()}
    \FOR {$l',v'$ \textbf{in} $m$.find\_most\_probable($K$)}
      \STATE $m' \leftarrow m$ with $v^{(t)}:=v'$
      \STATE $q'$.insert($l+l', m'$)
    \ENDFOR
  \ENDWHILE
  \STATE $q \leftarrow q'$
\ENDFOR
\STATE \textbf{return} $q$.pop()
\end{algorithmic}
\vskip 0.33em \small
$^\star$A \emph{min}-priority queue of fixed capacity $w$ maintains (at most) the $w$ \emph{highest} values at all times.
\end{algorithm}

%

When the output units $v^{(t)}_j,0\le j< N$ are conditionally independent given $\mathcal A^{(t)}$, such as for a regular RNN (eq.~\ref{eq:rnn_ce}),
it is possible to enumerate configurations by decreasing likelihood using a dynamic programming approach (Algorithm~\ref{alg-rnnmost}).
This very efficient algorithm in $O(K \log K + N\log N)$ is based on linearly growing priority queues, where $K$ need not be specified in advance.
Since inference is usually the bottleneck of the computation, this optimization makes it possible to use much higher beam widths $w$ with unbounded branching factors for RNNs.

\begin{algorithm}
\caption{\sc Independent outputs inference}
Enumerate the $K$ most probable configurations of $N$ independent Bernoulli random variables with parameters $0<p_i<1$.

\begin{algorithmic}[1] \label{alg-rnnmost} 
\STATE $v_0 \leftarrow \{i: p_i \ge \sfrac12\}$
\STATE $l_0 \leftarrow \sum_i \log (\max(p_i,1-p_i))$
\STATE \textbf{yield} $l_0$, $v_0$
\STATE $L_i \leftarrow |\log\frac{p_i}{1-p_i}|$
\STATE sort $L$, store corresponding permutation $R$
\STATE $q\leftarrow$ min-priority queue
\STATE $q$.insert($L_0$, $\{0\}$)
\WHILE {$l,v \leftarrow q$.pop()}
  \STATE \textbf{yield} $l_0-l$, $v_0 \triangle R[v]$~$^\star$
  \STATE $i \leftarrow \max(v)$
  \IF {$i+1 < N$}
    \STATE $q$.insert($l + L_{i+1}$, $v \cup \{i+1\}$)
    \STATE $q$.insert($l + L_{i+1} -L_i$, $v \cup \{i+1\} \setminus \{i\}$)
  \ENDIF
\ENDWHILE
\end{algorithmic}
\vskip 0.33em \small
$^\star A\triangle B \equiv (A\cup B)\setminus(A\cap B)$ denotes the symmetric difference of two sets.
$R[v]$ indicates the $R$-permutation of indices in the set $v$.
\end{algorithm}

A pathological condition that sometimes occurs with beam search over long sequences ($T\gg200$) is the exponential duplication of highly likely quasi-identical paths differing only at a few time steps, that quickly saturate beam width with essentially useless variations.
Several strategies have been tried with moderate success in those cases, such as committing to the most likely path every $M$ time steps (\emph{periodic restarts}~\cite{richter2010joy}), pruning similar paths, or pruning paths with identical $\tau$ previous time steps (the \emph{local assumption}), where $\tau$ is a maximal time lag that the chosen architecture can reasonably describe (e.g. $\tau\simeq200$ for RNNs trained with HF).
It is also possible to initialize the search with Algorithm~\ref{alg-beamsearch} then backtrack at each node iteratively, resulting in an anytime algorithm~\cite{zhou2005beam}.

\vspace{-1pt}
\section{Experiments} \label{sec:exp}
\vspace{-2pt}

In the following experiments, the acoustic input $x^{(t)}$ is constituted of powerful DBN-based learned representations~\cite{nam2011}.
The magnitude spectrogram is first computed by the short-term Fourier transform using a 128~ms sliding Blackman window truncated at 6~kHz, normalized and cube root compressed to reduce the dynamic range.
We apply PCA whitening to retain 99\% of the training data variance, yielding roughly $30\textrm{--}70\%$ dimensionality reduction.
A DBN is then constructed by greedy layer-wise stacking of sparse RBMs trained in an unsupervised way to model the previous hidden layer expectation ($v^{l+1}\equiv \mathbb E[h^l|v^l]$)~\cite{Bengio-2009}.
The whole network is finally finetuned with respect to a supervised criterion (e.g. eq.~\ref{eq:rnn_ce}) and the last layer is then used as our input $x^{(t)}$ for the spectrogram frame at time $t$.

We evaluate our method on five datasets of varying complexity:
Piano-midi.de, Nottingham, MuseData and JSB chorales (see~\cite{boulanger2012modeling}) which are rendered from piano and orchestral instrument soundfonts, and Poliner \& Ellis \cite{poliner2007discriminative} that comprises synthesized sounds and real recordings.
We use frame-level accuracy~\cite{bay2009evaluation} for model evaluation.
%
Hyperparameters are selected by a random search~\cite{bergstra2012random} on predefined intervals to optimize validation set accuracy; final performance is reported on the test set.

\begin{table}
\centering
\begin{tabular}{|l|c|c|c|}
\hline
Dataset & HMM \cite{nam2011} & RNN-RBM \cite{boulanger2012modeling} & Proposed \\
\hline \hline
Piano-midi.de & 59.5\% & 60.8\% & 64.1\% \\
Nottingham  & 71.4\% & 77.1\% & 97.4\% \\
MuseData & 35.1\% & 44.7\% & 66.6\% \\
JSB Chorales & 72.0\% & 80.6\% & 91.7\% \\
\hline
\end{tabular}
\vskip -1pt
\caption{Frame-level transcription accuracy obtained on four datasets by the Nam et al. algorithm with HMM temporal smoothing~\cite{nam2011}, using the RNN-RBM musical language model~\cite{boulanger2012modeling}, or the proposed I/O RNN-NADE model.}
\label{tab:jointmodel}
\end{table}

Table~\ref{tab:jointmodel} compares the performance of the I/O RNN-RBM to the HMM baseline~\cite{nam2011} and the RNN-RBM hybrid approach~\cite{boulanger2012modeling} on four datasets.
Contrarily to the product of experts of \cite{boulanger2012modeling}, our model is jointly trained, which eliminates duplicate contributions to the energy function and the related increase in marginals temperature, and provides much better performance on all datasets, approximately halving the error rate in average over these datasets.

\begin{figure}
\centering
\includegraphics[width=0.9\columnwidth]{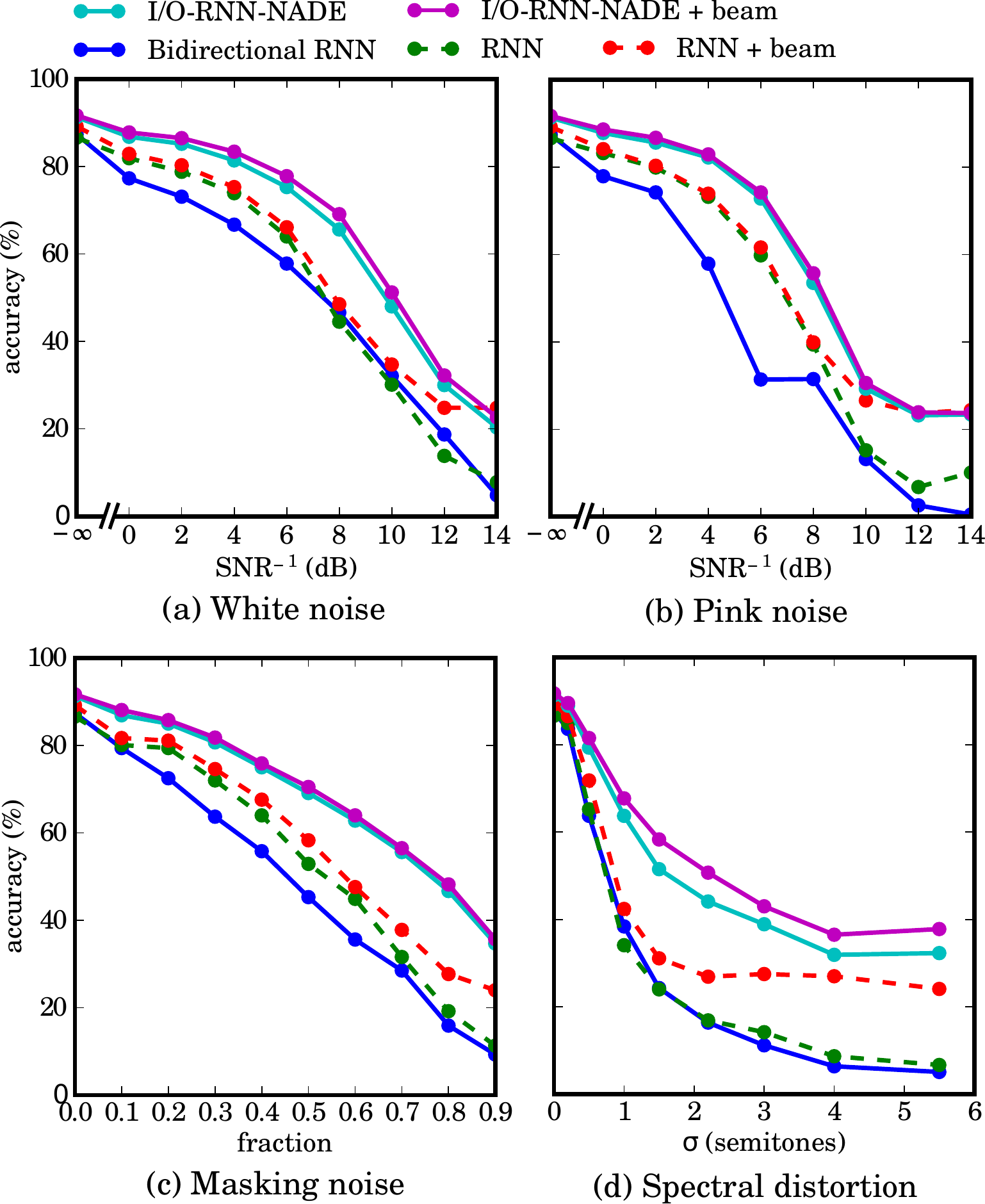}
\vskip -6pt
\caption{Robustness to different types of noise of various RNN-based models on the JSB chorales dataset.}
\label{fig:noise}
\end{figure}

\begin{figure}[!ht]
\centering
\includegraphics[width=0.9\columnwidth]{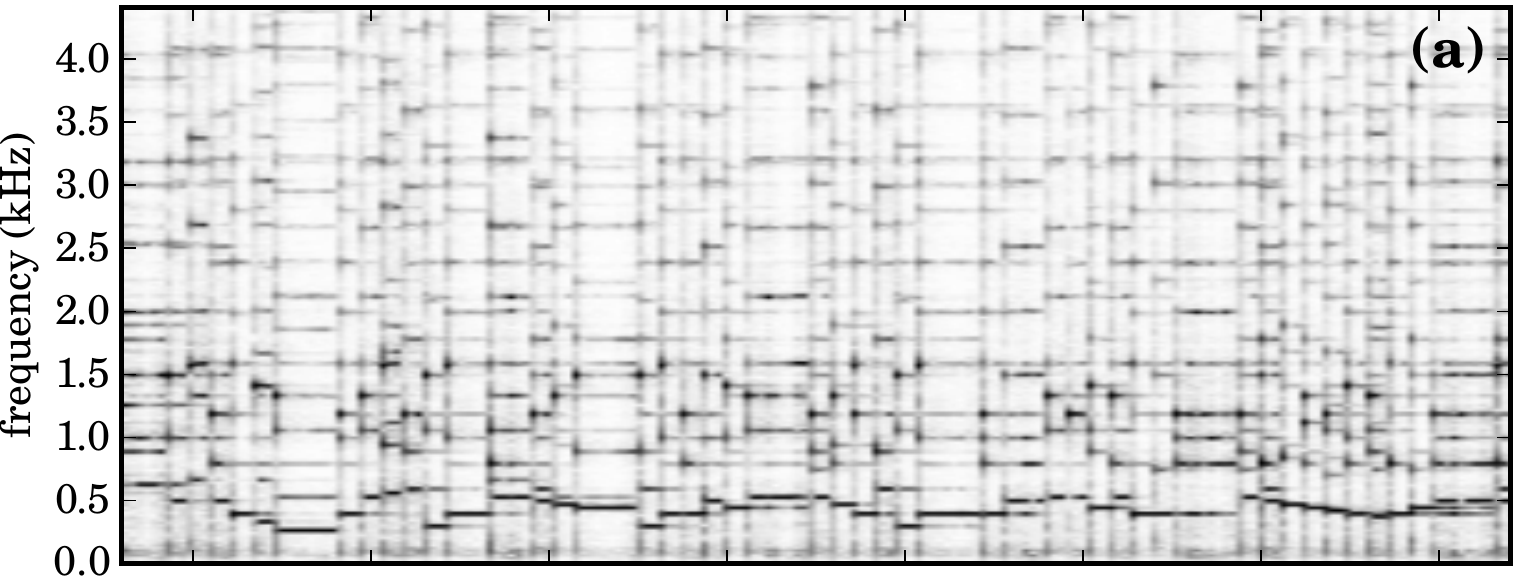} \vskip 0.05in
\includegraphics[width=0.9\columnwidth]{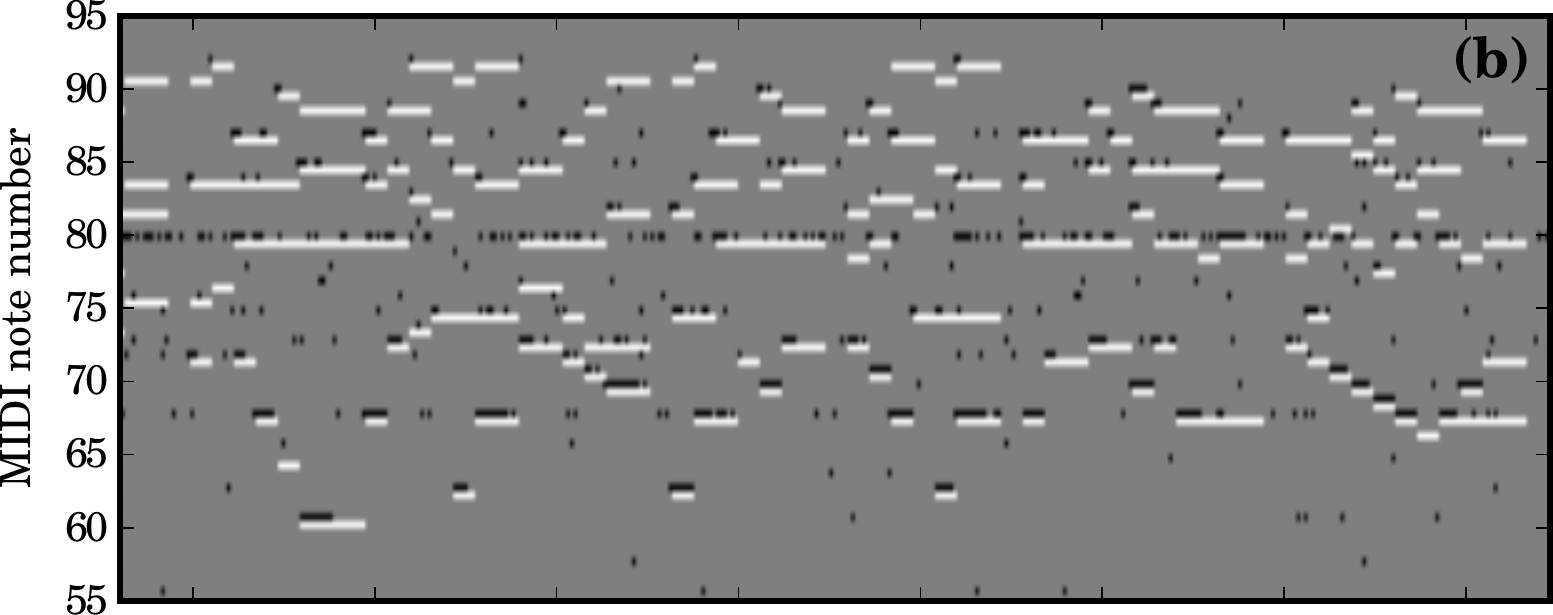} \vskip 0.05in
\includegraphics[width=0.9\columnwidth]{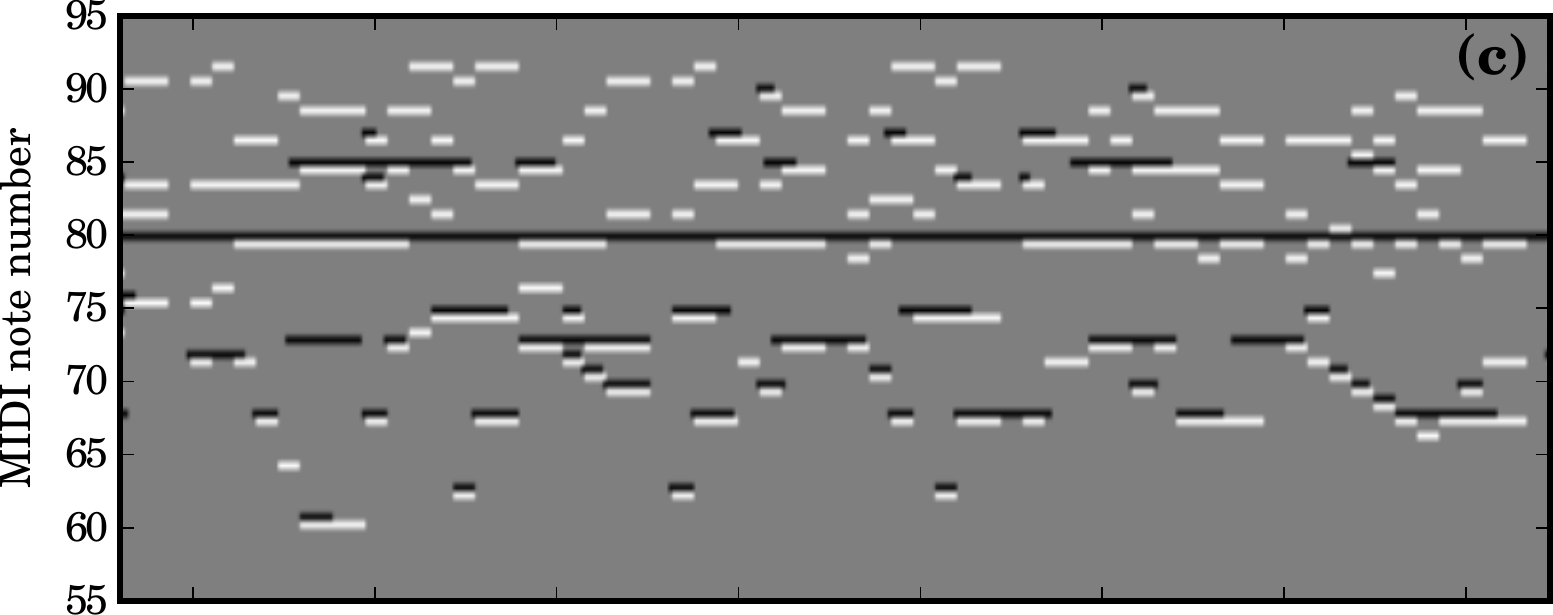} \vskip 0.05in
\includegraphics[width=0.9\columnwidth]{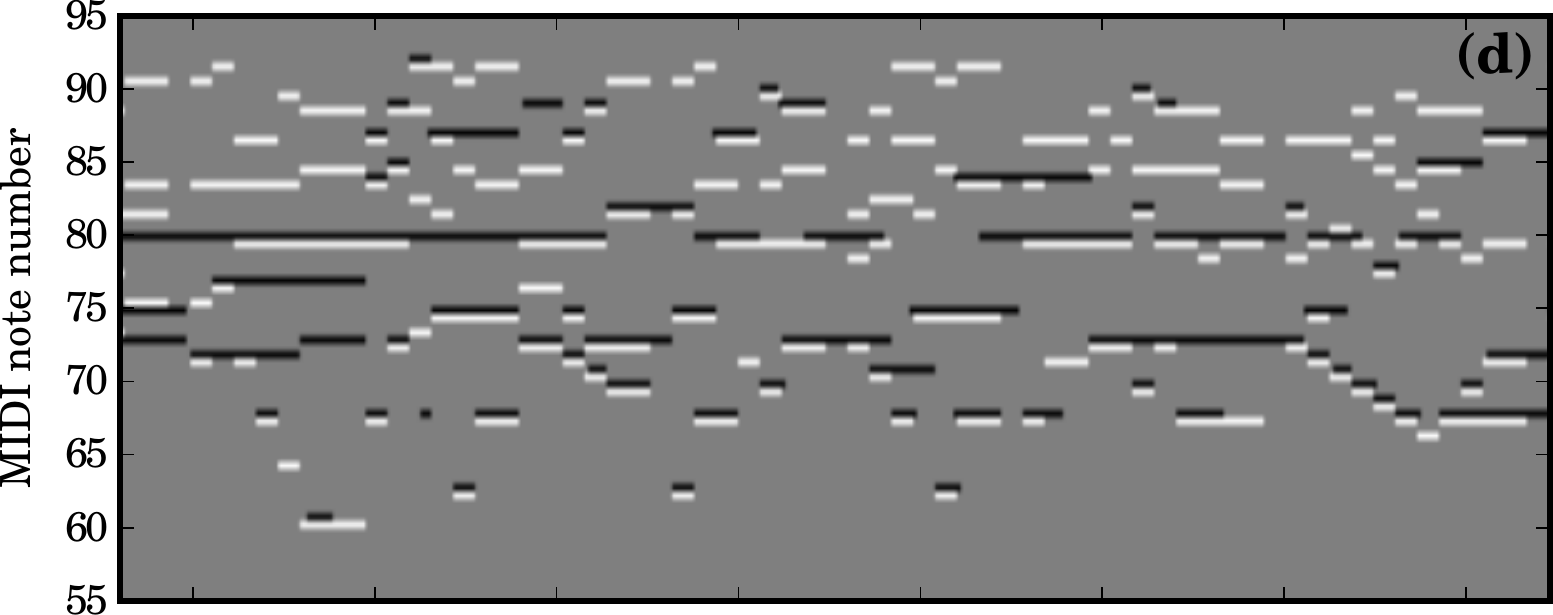} \vskip 0.05in
\includegraphics[width=0.9\columnwidth]{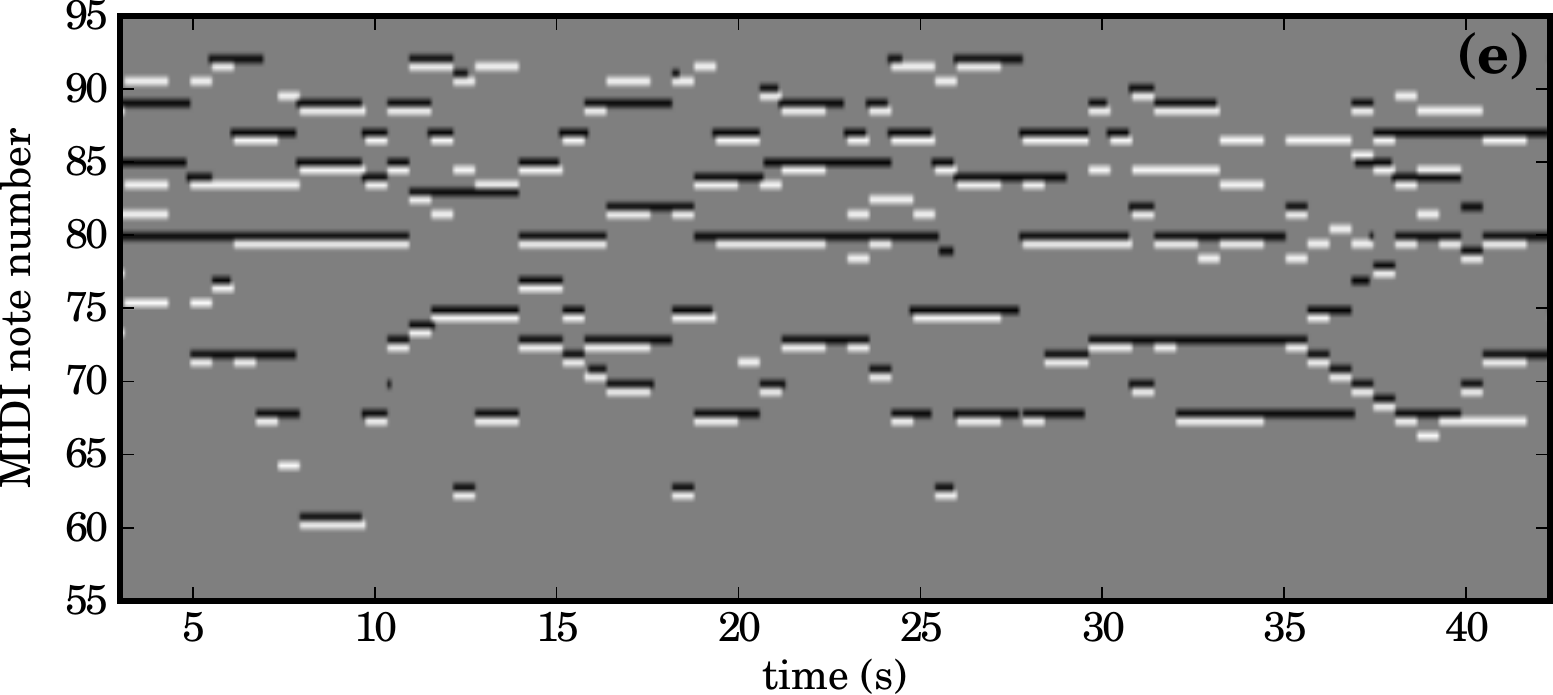}
\vskip -5pt
\caption{Demonstration of temporal smoothing on an excerpt of Bach's chorale \emph{Es ist genug} (BWV 60.5) with 6~dB pink noise. Figure shows (a) the raw magnitude spectrogram, and transcriptions by (b) a bidirectional RNN, (c) a bidirectional RNN with HMM post-processing, (d) an RNN with $v\rightarrow\hat h$ connections ($w=75$) and (e) I/O-RNN-NADE ($w=20$, $K=10$).
Predicted piano-rolls (black) are interleaved with the ground-truth (white) for comparison.}
\label{fig:demo_smoothing}
\end{figure} 

We now assess the robustness of our algorithm to different types of noise: white noise, pink noise, masking noise and spectral distortion.
In masking noise, parts of the signal of exponentially distributed length ($\mu=0.4$~s) are randomly destroyed \cite{vincent2008extracting};
spectral distortion consists in Gaussian pitch shifts of amplitude $\sigma$ \cite{palomaki2004techniques}.
The first two types are simplest because a network can recover from them by averaging neighboring spectrogram frames (e.g. Kalman smoothing), whereas the last two time-coherent types require higher-level musical understanding.
We compare a bidirectional RNN \cite{bock2012polyphonic} adapted for frame-level transcription, a regular RNN with $v\rightarrow\hat h$ connections ($w=2000$) and the I/O RNN-NADE ($w=50,K=10$).
Figure~\ref{fig:noise} illustrates the importance of temporal smoothing connections and the additional advantage provided by conditional distribution estimators. 
Beam search is responsible for a 0.5\% to 18\% increase in accuracy over a greedy search ($w=1$).

Figure~\ref{fig:demo_smoothing} shows transcribed piano-rolls for various RNNs on an excerpt of Bach's chorale \emph{Es ist genug} with 6~dB pink noise (Fig.~\ref{fig:demo_smoothing}(a)).
We observe that a bidirectional RNN is unable to perform temporal smoothing on its own (Fig.~\ref{fig:demo_smoothing}(b)), and that even a post-processed version (Fig.~\ref{fig:demo_smoothing}(c)) can be improved by our global search algorithm (Fig.~\ref{fig:demo_smoothing}(d)).
Our best model offers an even more musically plausible transcription (Fig.~\ref{fig:demo_smoothing}(e)).
Finally, we compare the transcription accuracy of common methods on the Poliner \& Ellis~\cite{poliner2007discriminative} dataset in Table~\ref{tab:trans_compare}, that highlights impressive performance.

\begin{table}
\centering
\begin{tabular}{|l|c|}
\hline
SONIC~\cite{marolt2004connectionist} & 39.6\%  \rule{0pt}{2.1ex} \\
Note events + HMM \cite{ryynanen2005polyphonic} & 46.6\% \\
Linear SVM \cite{poliner2007discriminative} & 67.7\% \\
DBN + SVM \cite{nam2011} & 72.5\% \\
BLSTM RNN \cite{bock2012polyphonic} & 75.2\% \\
AdaBoost cascade \cite{boogaart2009note} & 75.2\% \\
I/O-RNN-NADE & 79.1\% \\
\hline
\end{tabular}
\vskip -1pt
\caption{Frame-level accuracy of existing transcription methods on the Poliner \& Ellis dataset~\cite{poliner2007discriminative}.}
\label{tab:trans_compare}
\end{table}

\vspace{-1pt}
\section{Conclusions}
\vspace{-2pt}

We presented an input/output model for high-dimensional sequence transduction in the context of polyphonic music transcription.
Our model can learn basic musical properties such as temporal continuity, harmony and rhythm, and efficiently search for the most musically plausible transcriptions when the audio signal is partially destroyed, distorted or temporarily inaudible.
Conditional distribution estimators are important in this context to accurately describe the density of \emph{multiple} potential paths given the weakly discriminative audio.
This ability translates well to the transcription of ``clean" signals where instruments may still be buried and notes occluded due to interference, ambient noise or imperfect recording techniques.
Our algorithm approximately halves the error rate with respect to competing methods on five polyphonic datasets based on frame-level accuracy.
Qualitative testing also suggests that a more musically relevant metric would enhance the advantage of our model, since transcription errors often constitute reasonable alternatives.

\bibliographystyle{IEEEbib}
\bibliography{refs}

\end{document}